\pdfoutput=1
\documentclass{fairmeta}
\usepackage[utf8]{inputenc}
\usepackage{lmodern}
\usepackage{amsmath,amssymb,tabularx}
\setcitestyle{numbers,square,comma}

\title{KVUse: Reusable-KV User Memory for Efficient LLM Recommenders}
\author{Xin Guo}
\author{Qiang Zhang}
\author{Xiang Zhou}
\author{Arushi Rai}
\author{Siyu Lin}
\author{Yifan Wu}
\author{Xiang Shen}
\author{Lei Huang}
\author{Qunshu Zhang}
\author{Xiangjun Fan}
\author{Lizhu Zhang}
\affiliation{Meta Recommendation Systems (MRS)}
\correspondence{\email{xinguo@meta.com}}
\hypersetup{pdftitle={KVUse: Reusable-KV User Memory for Efficient LLM Recommenders},pdfauthor={Xin Guo, Qiang Zhang, Xiang Zhou, Arushi Rai, Siyu Lin, Yifan Wu, Xiang Shen, Lei Huang, Qunshu Zhang, Xiangjun Fan, Lizhu Zhang}}
\abstract{Memory-augmented recommenders built on large language models (LLMs) maintain a natural-language profile of each user. These profiles are updated by a user profile update path that revises the profile as new interactions arrive and are consumed by downstream task paths like ranking. Because these user profiles are stored as text in the user memory, every ranking LLM call that consumes them must splice them in the prompt and calculate the KV cache of them during the prefill phase. Given these ranking LLM calls are prefill heavy (long prompt but short decoded output) and repeatedly called, these repeated prefill of user profile text dominates the serving. Prefix caching cannot remove it, because these user profiles are preceded by different text in different calls. We present KVUse, which stores the user profiles as reusable key-value (KV) caches instead of text. KVUse trains lightweight LoRA writers by self-distillation, so that the user profile update call produces the reusable KV of the user profile while decoding it, and the LLM can read these reusable KV at new positions and behind new context without recomputation. KVUse also has a customized KV scheduler that uploads the stored reusable KV of the next batch while the current batch is computed, hiding the KV transfer from host memory behind GPU computation. On four InstructRec datasets with Llama-3.1-8B-Instruct, KVUse shows no statistically significant difference in ranking quality from full prefill over continuous memory update rounds. Under saturated load, KVUse raises the throughput of the downstream scoring path by 4.0\ensuremath{\times}--4.5\ensuremath{\times} and of the profile update path by 1.4\ensuremath{\times}--1.9\ensuremath{\times}.
}
\begin{document}
\maketitle
\FloatBarrier

\section{Introduction}

Large language models (LLMs) are turning recommender systems from static retrieval and ranking pipelines into systems that reason about users and items in natural language. This shift reaches the recommenders behind large Web platforms, such as e-commerce, book-review, and local-business sites, which personalize what every visiting user sees. Prompted with a user's history, an LLM can rank candidate items zero-shot \citep{ref6}, but such a ranker keeps no state between requests. Memory-augmented agentic recommenders \citep{ref16} close this gap by maintaining memory about users that is refined over time, through reflection on feedback and through signals shared across users and items \citep{ref2,ref11,ref12,ref15,ref17,ref25}.

A typical memory-augmented recommender consists of two paths (Figure 1). The user memory update path continuously collects the user's new engagement activities and consolidates this new information with the user's existing profile. The downstream task path reads the latest user profile and performs various downstream actions based on it, such as ranking the items that best match the user's interests. Both paths involve one or more LLM calls; Appendix A show example prompts.

The user profile these systems keep is interpretable, composed with arbitrary downstream prompts, and improves as the user interacts. However, it also introduces serving footprints that clash with the strict latency and capacity budgets of high-traffic recommenders. Table 1 gives a breakdown of latency for both paths (i.e. user profile update path and downstream scoring path) involved in a typical memory-augmented recommender system. From this table, we can see that downstream scoring path calls have much larger volume as compared with user profile update path, the frequency of which is more controllable and also less latency sensitive given it is an offline async job. We can also observe that the downstream scoring path calls are prefill-heavy and decode-light (at saturated throughput, prefill takes about 86\% of their engine time) while the user profile update path calls are decode-heavy (about 61\% of their engine time is decode). Taking both of the above two observations into consideration, we can see the most effective approach to improve the throughput of the system is to increase the efficiency of prefill steps for the downstream scoring path, which has larger traffic and is more latency sensitive.

\par\addvspace{12pt}
\noindent\begin{minipage}{\linewidth}
\centering
\captionof{table}{Traffic and engine work for the two paths of a memory-augmented recommender. Each percentage is the share of engine work that a call spends on prefill or on decode under saturated serving with full prefill.}\label{tab:1}
\begingroup
\fontsize{8.8}{11.0}\selectfont
\setlength{\tabcolsep}{3pt}
\renewcommand{\arraystretch}{1.18}
\begin{tabularx}{\linewidth}{@{}>{\hsize=0.8000\hsize\linewidth=\hsize\raggedright\arraybackslash}X>{\hsize=1.2000\hsize\linewidth=\hsize\raggedright\arraybackslash}X>{\hsize=1.0000\hsize\linewidth=\hsize\raggedright\arraybackslash}X>{\hsize=1.0000\hsize\linewidth=\hsize\raggedright\arraybackslash}X@{}}
\toprule
\textbf{Path} & \textbf{Calling frequency} & \textbf{Prefill share of engine work (saturated)} & \textbf{Decode share of engine work (saturated)} \\
\midrule
User profile update path & Every k (e.g., 10) new activities, async (e.g., daily/weekly) & 39\% (500 tokens) & 61\% (238 tokens) \\
Downstream scoring path & Upon user visits (e.g. 10 visits/day per DAU) & 86\% (513 tokens) & 14\% (9 tokens) \\
\bottomrule
\end{tabularx}
\endgroup
\end{minipage}
\par\addvspace{10pt}

The major factors that affect the prefill latency of downstream scoring calls are:

\begin{itemize}

\item \textbf{User profiles are recomputed for the same user}. A user profile is first computed as the output of a user profile update call. In the next update call for the same user, once new activities have been collected, the same profile text is spliced into the “Current profile” section of the user profile update prompt (Figure 4), as the base to be consolidated with the new activities into a new version of the profile. The same profile is also spliced into the “User Preferences” section of the downstream scoring prompt (Figure 5), where it represents the user's interests when the model scores a candidate item. In both cases, and especially in the second, which sits on the critical path of the downstream scoring path's prefill, the same profile text is spliced into multiple calls, and its KV is computed again in each of them.

\item \textbf{Item descriptions are recomputed across calls}. The same redundancy applies to item descriptions, which appear in the “New activity” section of the user profile update prompt and in the “Items to Score” section of the downstream scoring prompt.

\item \textbf{Prefix caching recovers only a small part of it} \citep{ref10,ref28}. A prefix cache can reuse the KV of a token only if the prompt up to that token is identical to one the engine has already processed. In a memory-augmented recommender, however, the text that precedes a user profile or an item description generally differs from call to call. For example, the profile in the user profile update prompt follows a system prompt that instructs the model to update the profile from the new activities and the existing profile, while the profile in the downstream scoring prompt follows a system prompt that instructs the model to score how well a candidate item matches the profile.

\end{itemize}

To address these gaps, we propose \textbf{KVUse} (Figure 2), a scalable memory-augmented recommender system that avoids this redundant computation by storing and reusing the KV cache of the memory instead of its text. Specifically, inspired by SemPIC \citep{ref21}, KVUse (1) trains a reusable-KV writer that lets the user profile update call output reusable KV for the profile it generates, which downstream scoring calls can splice in directly without recomputation. The writer is a LoRA adapter on the backbone LLM, trained to mitigate the KV reuse error. In addition, to reduce the overhead of uploading reusable KV from host memory to the GPU, KVUse (2) builds a KV scheduler that coordinates the scheduling of calls so that the KV upload for the next batch is hidden within the GPU computation of the current batch.

We evaluate KVUse on four datasets: Amazon Books, Goodreads, Amazon Movies \& TV, and Yelp. Under high concurrency, which is the regime closest to production serving, KVUse raises the throughput of the downstream scoring path by 4.0\ensuremath{\times}--4.5\ensuremath{\times} while maintaining the quality of the generated user profiles and of the downstream ranking. Our contributions are as follows:

\begin{itemize}

\item We identify the main factor behind the inference cost of memory-augmented recommenders, which are the repeated prefill of user and item information, and quantify its impact.

\item We propose \textbf{KVUse}, a KV-cache-based user memory system that removes this recomputation and raises the throughput of the downstream scoring path by 4.0\ensuremath{\times}--4.5\ensuremath{\times} while maintaining the quality of both the user profiles and the ranking.

\item We design a KV scheduler that hides the upload of reusable KV entirely within GPU computation without adding latency.

\end{itemize}

\begin{figure}[!htbp]
\centering
\includegraphics[width=\linewidth]{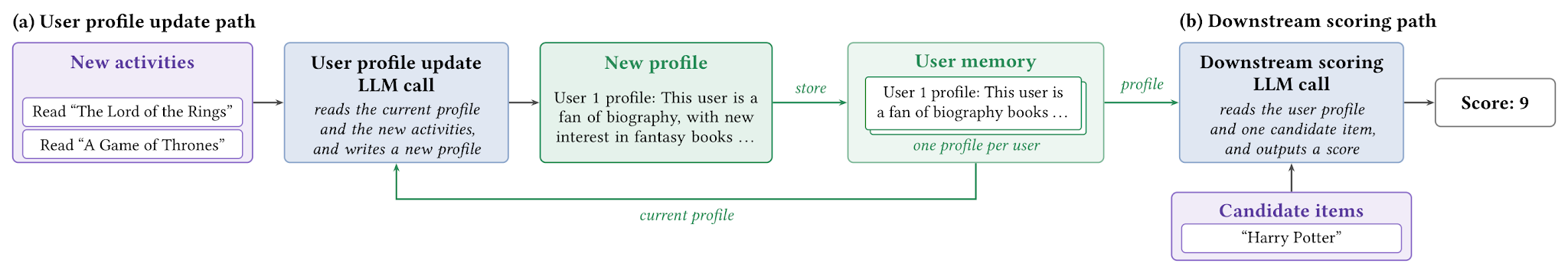}
\caption{The overall workflow of the memory-augmented recommender: (a) the user profile update path and (b) the downstream scoring path. Appendix A shows the prompts and responses of the two calls.}\label{fig:1}
\end{figure}

\begin{figure}[!htbp]
\centering
\includegraphics[width=\linewidth]{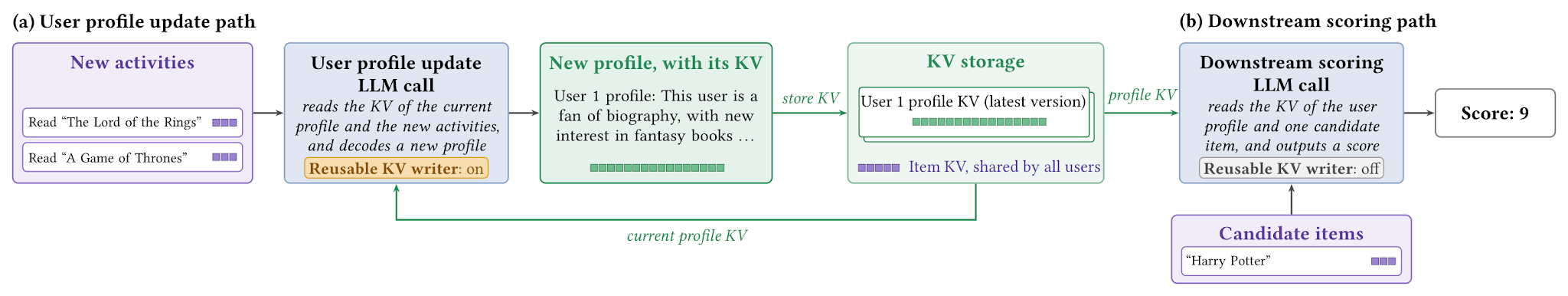}
\caption{The overall workflow of KVUse: (a) the user profile update path and (b) the downstream scoring path. Compared with Figure 1, the user memory is replaced by a KV storage that keeps the reusable KV of the latest user profiles and of the items, and both LLM calls read this KV instead of prefilling the corresponding text. A reusable KV writer is attached to the LLM: it is on while the update call decodes the new profile, so that the KV computed for the new profile can be stored and reused, and it is off in the scoring call.}\label{fig:2}
\end{figure}

\FloatBarrier

\section{Related Work}

\subsection{Memory-augmented LLM recommendation}

Using Large language models (LLMs) for recommendation tasks has attracted growing attention. Given a user's interaction history in the prompt, an LLM can rank candidate items without task-specific training \citep{ref6}. Recently, the idea of an agentic recommendation system has made further steps by adding planning, tool use, and memory modules \citep{ref16} on top of LLM base models. Particularly, the memory module clearly separates such agentic recommendation system from a zero-shot ranker: a zero-shot  ranker rebuilds its prompt from the raw history for every request, whereas a memory-augmented recommender maintains a representation of the user that persists across requests and is refined as new feedback arrives. Many memory-augmented recommenders keep a separate memory for each user, typically natural-language profiles that an LLM either revises from the user's own feedback \citep{ref22} or generates from representative sub-sequences of a long behavior history \citep{ref18}; RecGPT-V3 \citep{ref27} brings such per-user memory to an industrial recommender. Others share memory across users and items, so that feedback from one user can also inform recommendations for others: AgentCF \citep{ref25} treats users and items alike as agents whose memories are updated jointly whenever a simulated choice disagrees with the real interaction, and MemRec \citep{ref2} propagates new insights over a memory graph that links users with items.

The system closest to ours is MARS \citep{ref17}, which keeps a three-tier memory for each user, ranging from raw interaction events through fine-grained preference statements in natural language to a profile paragraph that summarizes those statements. As interactions accumulate, an LLM planner periodically revises the preference tier by adding, strengthening, weakening, merging, or deleting statements, and the profile is regenerated automatically after every 5 such changes. At recommendation time, an LLM ranker scores the candidate items conditioned on the profile and the user's most recent events. The memory loop studied in this paper is a simplified version of this pipeline, and our experiments use the same four InstructRec domains on which MARS is evaluated.

\subsection{KV cache reuse for LLM inference}

Reusing KV caches across requests is a standard way to reduce prefill computation, and existing methods differ in whether a cached segment must keep the positions at which it was computed \citep{ref13}.

\textbf{Position-dependent caching.} Prefix caching in vLLM \citep{ref10} and SGLang \citep{ref28} shares the KV of an identical prompt prefix, and Prompt Cache \citep{ref4} extends reuse to modules declared in a prompt schema; in both, a segment is reused only at the position where it was computed.

\textbf{Position-independent caching (PIC).} Formalized by EPIC \citep{ref8}, PIC lets a cached segment be reused at any position, which introduces two errors: its keys still encode the old positions, and its KV lacks attention to the new context that precedes it. The positional error can be corrected exactly by re-rotating the keys, so methods differ in how they recover the missing cross-segment attention.

Training-free methods recompute a small subset of the reused tokens on every request. CacheBlend \citep{ref24} selects the tokens whose KV deviates most from a full prefill; EPIC \citep{ref8} recomputes the first few tokens of every chunk except the first, because these tokens would otherwise act as spurious attention sinks; ProphetKV \citep{ref20} recomputes the tokens that the user's query attends to most, and RelayCaching \citep{ref3} applies selective recomputation to the KV that an upstream agent produces while decoding. All of them add recomputation to every request that reads the cache.

Training-based methods instead prepare the model so that independently encoded segments can be concatenated without recomputation. Block-Attention \citep{ref14} and KVLink \citep{ref23} fine-tune the LLM itself, which then must replace the original model in serving, whereas KV Packet \citep{ref1} and SemPIC \citep{ref21} keep the reader frozen and train only the writing side, with a few soft tokens around each cached document or a LoRA writer that encodes each document offline. KVUse follows the frozen-reader approach but applies it to generated content: like RelayCaching, it takes the user profile KV from the decoding step of an upstream call, and this KV serves both downstream scoring and the next memory update.

\subsection{KV cache reuse in recommendation}

To the best of our knowledge, no prior work reuses KV caches in a memory-augmented recommender. The closest use of KV reuse in recommendation is RcLLM \citep{ref26}, which reuses the KV of candidate items and of clustered user engagement histories for LLM ranking. Its user histories, which RcLLM gives as the text of past reviews, are approximated and therefore not fully personalized: RcLLM embeds each history token, groups the embeddings into a bounded set of semantic prototypes, and precomputes KV for each prototype; at serving time, each history token is replaced by the KV of its nearest prototype. In addition, RcLLM keeps no persistent per-user memory, so it cannot enjoy the benefits of a memory system: a compact profile that summarizes a long history, an understanding of the user that accumulates and is refined across interactions, and an interpretable representation that any downstream task can read. Finally, its KV reuse runs full attention in the first layer to select heavy-hitter tokens and recomputes them, together with a local window, in later layers, which adds computation to every request.

\FloatBarrier

\section{Method}

\subsection{Memory-augmented Recommender System}

We study a simplified MARS-style memory loop \citep{ref17}, which is consisted of two paths (Figure 1):

\begin{itemize}

\item \textbf{The user profile update path} receives the current profile and the user's newest interactions, and writes a revised profile. Figure 4 gives an example of the prompt and response for this call. The prompt is consisted of 4 parts: a consistent prefix for task instruction which is shared by all calls; a “Current profile” section to inject current user profile from the user memory; a “New activity” section to include all new items the user have newly engaged; and an ending instruction to generate the output.

\item \textbf{The downstream scoring path} fetches the current user profile from user memory and is given one candidate item, and returns a relevance score. Figure 5 gives an example of the prompt and response for this call. The prompt consists of 4 parts: a consistent prefix for task instruction; a “User Preference” section to inject current user profile from the user memory; a “Item to Score” section to keep the information of the item candidates; and an ending instruction to generate the output.

\end{itemize}

\subsection{Insights and Motivations}

\textbf{Insight 1: Computational Redundancy in User Profile and Item Description}. As mentioned in the instruction part, the user profiles in the user memory storage, once generated, are reused in the user profile update LLM call and downstream scoring LLM calls. As a result, the KV cache of these user profile tokens are computed when they get generated for the first time and then recomputed multiple times when they are used as context in the prompts of future user profile update calls and downstream scoring calls (red colored in Figures 4 and 5). Ideally the KV cache of these user profiles should only be calculated once and get reused for future calls. This redundancy is similar for item description, which is static and known before inference, but the KV cache of these item descriptions get recomputed in various LLM calls (orange colored in Figures 4 and 5). This indicates that the KV cache of these item descriptions should be precomputed offline and get reused in future calls.

\textbf{Insight 2: Attention sparsity}. Figure 3 shows the attention heatmap for the prompt of a typical user profile update LLM call, which includes 3 major triangle chunks: the upper chunk is the positions for the system instruction, the middle chunk is the positions for the user profile, and the lower chunk is the positions for item description. These centralized chunks show that the majority of the attentions are within chunk attention while the cross-chunk attention are sparse, which indicates the possibility of KV cache reuse - by directly spice pre-computed reusable KV cache for each chunk, we are only using the within chunk attention in the prompt part and a sparse cross chunk attention can potentially reduce the error. One thing worth to be noted is that in the decoding part, the output token positions can still have the full attention to all previous chunks in the prompt.

\begin{figure}[!htbp]
\centering
\includegraphics[width=0.70\linewidth]{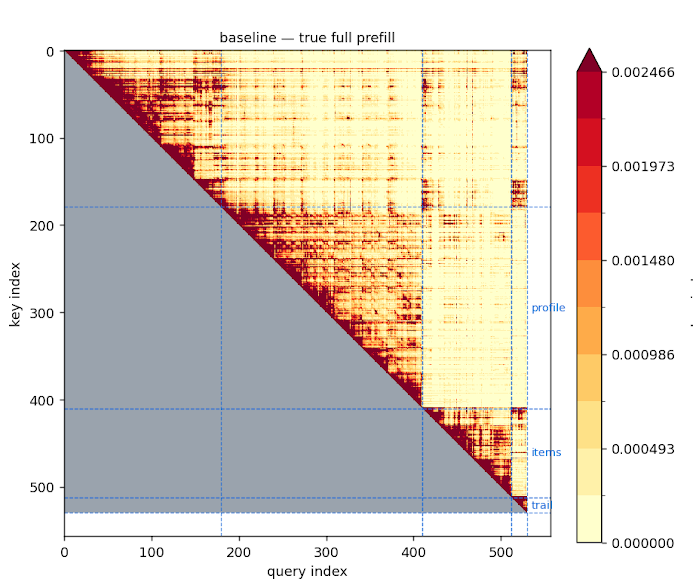}
\caption{Attention heatmap for a user profile update LLM call (full compute, no KV reuse).}\label{fig:3}
\end{figure}

\textbf{Insight 3: Natively reuse KV cache can bring significant ranking quality degradation. }Reusing the KV of a chunk that is not a shared prefix introduces two errors:

\begin{itemize}

\item \textbf{Position.} Rotary position embeddings \citep{ref19} bake the producer's positions into the cached keys, but this error can be removed exactly. RoPE encodes a token's position by rotating its key through angles proportional to that position, and rotations applied one after another add their angles. Rotating a stored key further by the angle that corresponds to the shift from its original position to its new one therefore yields exactly the key that the consumer would have computed. Values carry no positional encoding and are copied unchanged.

\item \textbf{Context.} Each token in the user profile chunk attends to all of its previous context and baked this context information into its KV cache. When these chunks get reused and spiced into another call, the context is changed and previous cached KV may not be accurate anymore. As can be seen from Table 3, natively re-using KV cache (only fixing the position embedding) can generally drop NDCG@10 by 20\%-30\%. This indicates that in order to generate a reusable KV cache without significantly ranking quality degradation, we need to modify the KV cache that can be compatible for various contexts.

\end{itemize}

\subsection{System Design}

In this paper, we propose KVUse, a scalable memory-augmented LLM recommendation system. As illustrated in Figure 2, the overall system consisted of user profile update paths and downstream scoring paths.

\textbf{Training time.}

During the training time, KVUse learns its reusable KV writers, which are the only trainable component of the system. Following SemPIC \citep{ref21}, we separate writing reusable KV from reading it. Both roles share the same pretrained LLM, whose parameters \emph{\ensuremath{\theta}} stay frozen throughout training. A writer attaches a LoRA adapter \emph{\ensuremath{\phi}\textsubscript{s}} to the query, key, value, and output projections of every layer, and the adapter is switched on only while the model produces the KV of the writer's span type \emph{s}. The reader is the vanilla LLM with every LoRA adapter switched off. We train one writer per span type, namely the user profile and the item descriptions. Training therefore alters only the KV placed in the KV storage, while the LLM that consumes this KV is left untouched.

The training data are produced by the vanilla LLM itself, so no human labels are needed. For each user in the training data, we first run the LLM calls with full prefill and without any reusable KV, which we call the teacher chain. For example, for the user profile update path, the first LLM call writes an initial profile from the user's earliest interactions, and three update calls then revise it, each with the user's next interactions, using the prompts of Figure 4 and greedy decoding of up to 640 tokens, including the end-of-sequence token. Each update call contributes a user profile update record, which pairs the prompt (Figure 4) with the profile the teacher wrote. For the user profile of every round, we also score the user's 10 candidate items with the scoring prompt (Figure 5), namely the held-out item the user actually engaged with and 9 negatives from the dataset's candidate list, and each of these calls contributes a scoring record that holds the teacher's score. We train the writers on these records with teacher forcing: the teacher's answer is appended to each prompt, so that at every answer position the student and the teacher predict the next token from the same preceding tokens, and a whole call takes a single forward pass instead of a decoding loop.

The student path reproduces serving for a complete training sample, comprising the following three kinds of calls: the profile update call, the scoring calls that read the profile this call writes, and the next update call. (1) Each item description is written by the item writer on its own, starting at position zero, so that its KV can be stored once and shared across users and prompts. (2) The update call reads two pieces of reusable KV: the current profile KV, produced the same way in the previous round, and the KV of the newly engaged items. Instead of generating, it is fed the teacher's new profile as its answer. Because the profile writer is enabled over these answer tokens, the KV rows they generated form the new profile KVs, which are exactly the KVs that the following update calls use when it decodes the same text at serving time. (3) The scoring prompts and the next update prompt are assembled from the system instruction, which the LLM computes as usual, followed by reusable KV pieces: the new profile KV, and the KV of the candidate item or of the next newly engaged items. Every piece is re-rotated to its position in the prompt. (4) The KV reader consumes every prompt and predicts the teacher's answer token by token. In the two update calls, the profile writer is enabled over the answer, which is the same as the serving time; in the scoring calls, every writer is disabled. (5) At every answer token, the student's distribution is compared with the teacher's, which the vanilla LLM computes from the complete prompt with full prefill, and these comparisons define the loss below.

Formally, for a prompt \emph{x} with teacher answer \emph{y} = (\emph{y}\textsubscript{1}, \ldots{}, \emph{y\textsubscript{T}}), the distillation loss of one call is the forward KL divergence between the teacher's and the student's next-token distributions, averaged over the answer tokens:

\begin{equation}\label{eq:1}
L_{\mathrm{KD}}(x,y)=\frac{\tau^2}{T}\sum_{t=1}^{T}\mathrm{KL}\!\left(p_{\theta}(\cdot\mid x,y_{<t})\,\Vert\,p_{\theta,\phi}(\cdot\mid\tilde{x}_{\phi},y_{<t})\right)
\end{equation}

Here \emph{x} is the prompt of one call, as a sequence of tokens, and \emph{y} is the teacher's answer to it: the profile written by an update call, or the score returned by a scoring call. \emph{T} is the number of answer tokens, including the end-of-sequence token, and \emph{y}\textsubscript{\textless{}}\emph{\textsubscript{t}} = (\emph{y}\textsubscript{1}, \ldots{}, \emph{y\textsubscript{t}}\textsubscript{\ensuremath{-}1}) denotes the answer tokens before position \emph{t}. \emph{p\textsubscript{\ensuremath{\theta}}}(\ensuremath{\cdot} | \emph{x}, \emph{y}\textsubscript{\textless{}}\emph{\textsubscript{t}}) is the teacher's distribution over the vocabulary for the \emph{t}-th answer token, which the vanilla LLM with parameters \emph{\ensuremath{\theta}} produces when it reads \emph{x} with full prefill, followed by \emph{y}\textsubscript{\textless{}}\emph{\textsubscript{t}}. \emph{\ensuremath{\tilde{x}}\textsubscript{\ensuremath{\phi}}} is the student's version of \emph{x}: it has the same tokens, but its profile and item spans are supplied as KV written by the writers with parameters \emph{\ensuremath{\phi}} and re-rotated to their positions. \emph{p\textsubscript{\ensuremath{\theta}}}\textsubscript{,}\emph{\textsubscript{\ensuremath{\phi}}}(\ensuremath{\cdot} | \emph{\ensuremath{\tilde{x}}\textsubscript{\ensuremath{\phi}}}, \emph{y}\textsubscript{\textless{}}\emph{\textsubscript{t}}) is the student's distribution for the same token, which the reader produces when it consumes \emph{\ensuremath{\tilde{x}}\textsubscript{\ensuremath{\phi}}} followed by \emph{y}\textsubscript{\textless{}}\emph{\textsubscript{t}}, with the profile writer on over \emph{y}\textsubscript{\textless{}}\emph{\textsubscript{t}} in an update call. KL(\emph{p} \ensuremath{\Vert} \emph{q}) = \ensuremath{\sum}\emph{\textsubscript{v}} \emph{p}(\emph{v}) log(\emph{p}(\emph{v}) / \emph{q}(\emph{v})) is the Kullback--Leibler divergence, summed over the vocabulary. Both distributions are computed from logits divided by the temperature \emph{\ensuremath{\tau}}, which we set to 1, and the factor \emph{\ensuremath{\tau}}\textsuperscript{2} keeps the gradient scale independent of \emph{\ensuremath{\tau}}. Dividing by \emph{T} puts long profiles and short scores on the same per-token scale. The loss of one training sample adds up its three kinds of calls:

\begin{equation}\label{eq:2}
\begin{aligned}L(\phi)={}&L_{\mathrm{KD}}(x_{1}^{\mathrm{upd}},y_{1}^{\mathrm{upd}})+\lambda\sum_{j=1}^{10}L_{\mathrm{KD}}(x_{j}^{\mathrm{score}},y_{j}^{\mathrm{score}})\\&+L_{\mathrm{KD}}(x_{2}^{\mathrm{upd}},y_{2}^{\mathrm{upd}}),\qquad\lambda=1\end{aligned}
\end{equation}

Here \emph{x}\textsubscript{1}\textsuperscript{upd} is the user profile update prompt and \emph{y}\textsubscript{1}\textsuperscript{upd} the teacher's new profile, whose KV rows in the student form the new profile KV. \emph{x\textsubscript{j}}\textsuperscript{score} is the scoring prompt of the \emph{j}-th of the 10 candidate items, which reads this profile KV, and \emph{y\textsubscript{j}}\textsuperscript{score} is the teacher's score for that item. \emph{x}\textsubscript{2}\textsuperscript{upd} is the next user profile update prompt, which reads the same profile KV, and \emph{y}\textsubscript{2}\textsuperscript{upd} is the teacher's next profile. We call the two update terms the update loss and the sum over the candidates the scoring loss; \emph{\ensuremath{\lambda}} weights the scoring loss against the update loss. The teacher's distributions are computed without gradients and \emph{\ensuremath{\theta}} is never updated, so minimizing \emph{L}(\emph{\ensuremath{\phi}}) changes only the writers. The writer KV is not detached before the reader consumes it, so the gradient reaches \emph{\ensuremath{\phi}} through the KV that the writers produce and, in the two update calls, through the answer tokens that are read with the profile writer on.

Training proceeds in two stages. Stage A trains each writer in a separate run, in which only that writer's span is supplied as writer KV, every other span is computed with full prefill, and the other writers are kept fixed. Stage B initializes all writers from their Stage A checkpoints and fine-tunes them jointly, with every reusable span supplied as writer KV and with both terms of Eq. (2), the update loss and the scoring loss, so that the writers adapt to one another.

\textbf{Inference time.}

During inference, the user profile update path for a single user is triggered by the profile update planner. We can configure the frequency of triggering the user profile update path by activities counting based approach (e.g. triggering on every N new activities) or time based approach (e.g. daily/weekly triggering). Once triggered, the profile update planner will fetch all of required materials in order to prepare the prompt for the user profile update calls: fetching the reusable KV cache for current user profile for this user and fetching the reusable KV cache for the descriptions of items newly engaged by this user from the KV storage. Once all materials get collected, they will be sent to the inference scheduler, which can coordinate the reusable KV cache upload and LLM inference across different batches.

In order to completely hide the latency of uploading reusable KV from host side to the GPU, the inference scheduler bundles pending LLM requests into batches and internally maintains a FIFO queue to keep the ordering across batches. Then the inference scheduler uses a side-stream to async uploads the reusable KV cache for next request batch via KV connector while the GPUs are still busy computing the current request batch. With this KV uploading and computing parallelism, we can completely hide the latency of KV uploading under the GPU computing latency without introducing extra latency.

Once the LLM serving engine (e.g. vLLM) received the request, in the prefill step, it will use vanilla LLM model to directly read the reusable KV caches for the user profile and items which are previously uploaded to HBM by inference scheduler and only compute very small amount of suffix tokens that are not covered by the reusable KV caches. It is worth to be noted that the system instruction chunk at the beginning of each call can be directly fetched from native prefix cache provided by the LLM serving engines like vLLM or SGLang. During the decoding step, the LLM model will turn on a reusable KV writer, which is a trained LoRA, to output KVs of the new user profile version along with the user profile text tokens. These generated KVs and corresponding text tokens for the new user profile version are optimized for minimizing the performance degradation of downstream scoring tasks, so they are expected to be different from the output from vanilla LLM’s output with the reusable KV writer disabled. The detailed comparison of this difference can be found in the Experiments section below. In the end, these generated KVs of the new user profile will be stored in the KV storage for future usage.

The downstream scoring path also follows the similar workflow. This path is also triggered by the scoring planner, which can be configured to be offline triggered for offline scoring tasks or online triggered whenever a live ranking request comes in to the planner. Once triggered, the scoring planner will also fetch the reusable KV cache of user profile and items from the KV storage. For online use cases, we may want to add a cache layer (e.g. memcached) on top of KV storage to speed up the reusable KV fetching process. The scoring planner will also send the requests along with the fetched reusable KV cache to the inference scheduler for KV upload and request calculation. The LLM model will use the reusable KV cache in the prefill steps and directly output the matching scores in the decoding steps. Through this LLM call, the reusable KV writer is disabled.

\FloatBarrier

\section{Experiments}

\subsection{Experiment Setup}

\textbf{Datasets.} Our experiments use four domains of the InstructRec benchmark released with iAgent \citep{ref22}: Amazon Books, Goodreads, Amazon Movies \& TV, and Yelp. They are built from book, movie, and business reviews and differ in size and density (Table 2). We use only the users' interaction histories, not the natural-language instructions attached to them. Amazon Books is our main dataset: its users are split by a hash of their id into disjoint training and evaluation sets of 1,500 and 200 users, and quality is reported on the 160 evaluation users whose histories are long enough for four memory rounds. On Goodreads, Amazon Movies \& TV, and Yelp, quality is reported on 800, 711, and 484 evaluation users.

\par\addvspace{12pt}
\noindent\begin{minipage}{\linewidth}
\centering
\captionof{table}{Statistics of the four datasets \citep{ref17} and the number of users on which ranking quality is evaluated.}\label{tab:2}
\begingroup
\fontsize{8.2}{10.2}\selectfont
\setlength{\tabcolsep}{3pt}
\renewcommand{\arraystretch}{1.18}
\begin{tabularx}{\linewidth}{@{}>{\hsize=1.1842\hsize\linewidth=\hsize\raggedright\arraybackslash}X>{\hsize=0.8772\hsize\linewidth=\hsize\raggedright\arraybackslash}X>{\hsize=0.8772\hsize\linewidth=\hsize\raggedright\arraybackslash}X>{\hsize=1.0965\hsize\linewidth=\hsize\raggedright\arraybackslash}X>{\hsize=0.9649\hsize\linewidth=\hsize\raggedright\arraybackslash}X@{}}
\toprule
\textbf{Dataset} & \textbf{Users} & \textbf{Items} & \textbf{Interactions} & \textbf{Evaluated users} \\
\midrule
Amazon Books & 7,377 & 120,925 & 207,759 & 160 \\
Goodreads & 11,734 & 57,364 & 618,330 & 800 \\
Amazon Movies \& TV & 5,649 & 28,987 & 79,737 & 711 \\
Yelp & 2,950 & 31,636 & 63,142 & 484 \\
\bottomrule
\end{tabularx}
\endgroup
\end{minipage}
\par\addvspace{10pt}

\textbf{Writer training.} Each writer is a LoRA adapter \citep{ref7} with rank 8 and alpha 16 on the query, key, value, and output projections of every layer of Llama-3.1-8B-Instruct \citep{ref5}, adding 6.8M trainable parameters. Writers are trained separately for each dataset; on Amazon Books, they are trained on the teacher chains of the 1,500 training users, which contain 4,500 update records and 60,000 scoring records. Both stages use AdamW with 4 training samples per step and a budget of 2,400 steps, over which the learning rate decays linearly to one tenth of its initial value, 5e-4 in Stage A and 6e-5 in Stage B.

\textbf{Baselines.} All methods use Llama-3.1-8B-Instruct with greedy decoding and the prompts of Appendix A. We compare KVUse with:

\begin{itemize}

\item \textbf{Full prefill:} the conventional memory-augmented recommender, which stores each profile as text and prefills every prompt in full. When serving, it uses vLLM's automatic prefix caching, so the scoring prompts of a slate share the prefill of their common prefix, including the profile.

\item \textbf{Naive KV reuse:} KVUse with zero-initialized writers. Every stored span therefore holds the KV that the vanilla LLM computed where the span first appeared, and only its positions are corrected by exact re-rotation. This baseline isolates the context error (Section 3.2).

\end{itemize}

\textbf{Evaluation protocol.} We adopt the leave-one-out protocol of prior work \citep{ref2,ref22}. The latest interaction of each user is held out as the ground-truth item and mixed with 9 negatives drawn at random, giving 10 candidates per user; the candidates stay the same in every memory round.

\textbf{Metrics.} For accuracy, we report NDCG@10 as the primary ranking metric, and NDCG@5, HR@1, and HR@5 as secondary metrics. For efficiency, we report saturated throughput and latency separately for the two paths, because they are invoked at very different rates (Table 1). On the update path, throughput is the number of update requests completed per second; on the scoring path, it is the number of users served per second, where a user counts as served once all 10 of their candidates have been scored. Latency is the mean engine step time under the same load. Finally, we measure profile fidelity, that is, whether the profiles that KVUse writes describe the user's interests and engagement habits as accurately as the profiles that the vanilla LLM writes with full prefill, using an LLM judge (Section 4.3).

\textbf{Serving setup.} We serve KVUse and full prefill with vLLM 0.26 \citep{ref10}. Requests are submitted ahead of the engine, so that its waiting queue never runs empty and every engine step is filled up to vLLM's own scheduling limits. Throughput is counted over a fixed window in the middle of the run, which excludes warm-up and drain, and we report the median of three runs. Both arms run the same model on the same AMD MI300X GPU and engine configuration, keep automatic prefix caching enabled, serve the same pool of requests, and decode the same average number of output tokens; they differ only in where the KV of the profile and item spans comes from. Full prefill computes the KV of every prompt token that the prefix cache does not cover, whereas KVUse loads the stored KV of the profile and item spans from host memory through a vLLM KV connector, computes only the remaining tokens, and uses the KV scheduler to upload the KV of the next batch in advance.

\subsection{Downstream Ranking Quality over the Memory Chain}

\par\addvspace{12pt}
\noindent\begin{minipage}{\linewidth}
\centering
\captionof{table}{NDCG@10 per memory update round on Amazon Books (160 held-out users). \ensuremath{\Delta} is KVUse minus full prefill. Appendix B gives NDCG@5, HR@1, and HR@5.}\label{tab:3}
\begingroup
\fontsize{8.2}{10.2}\selectfont
\setlength{\tabcolsep}{3pt}
\renewcommand{\arraystretch}{1.18}
\begin{tabularx}{\linewidth}{@{}>{\hsize=1.8000\hsize\linewidth=\hsize\raggedright\arraybackslash}X>{\hsize=0.8000\hsize\linewidth=\hsize\raggedright\arraybackslash}X>{\hsize=0.8000\hsize\linewidth=\hsize\raggedright\arraybackslash}X>{\hsize=0.8000\hsize\linewidth=\hsize\raggedright\arraybackslash}X>{\hsize=0.8000\hsize\linewidth=\hsize\raggedright\arraybackslash}X@{}}
\toprule
\textbf{Arm} & \textbf{Round 0} & \textbf{Round 1} & \textbf{Round 2} & \textbf{Round 3} \\
\midrule
Full prefill & 0.6152 & 0.6128 & 0.6306 & 0.6255 \\
Naive KV reuse (relocation only)  & 0.4564 & 0.4491 & 0.4658 & 0.4559 \\
\textbf{KVUse}  & \textbf{0.5957} & \textbf{0.6248} & \textbf{0.6454} & \textbf{0.6502} \\
\ensuremath{\Delta} (KVUse \ensuremath{-} full prefill) & \ensuremath{-}0.0195 & +0.0119 & +0.0148 & +0.0246 \\
\bottomrule
\end{tabularx}
\endgroup
\end{minipage}
\par\addvspace{10pt}

Table 3 compares KVUse with full prefill and naive KV reuse over the four memory rounds on Amazon Books, and Table 4 extends the comparison to all four datasets. Appendix B.1 further compares KVUse with two read-time repair methods, CacheBlend and EPIC.

\par\addvspace{12pt}
\noindent\begin{minipage}{\linewidth}
\centering
\captionof{table}{Ranking quality averaged over four profile update rounds. Columns show full-prefill NDCG@10, KVUse minus full prefill (95\% confidence intervals over users), and the NDCG@10 change of naive KV reuse. On Goodreads, Amazon Movies \& TV, and Yelp, naive reuse is evaluated on full-prefill profiles with KV reused only for scoring.}\label{tab:4}
\begingroup
\fontsize{7.0}{8.8}\selectfont
\setlength{\tabcolsep}{3pt}
\renewcommand{\arraystretch}{1.18}
\begin{tabularx}{\linewidth}{@{}>{\hsize=1.0000\hsize\linewidth=\hsize\raggedright\arraybackslash}X>{\hsize=1.0000\hsize\linewidth=\hsize\raggedright\arraybackslash}X>{\hsize=1.0000\hsize\linewidth=\hsize\raggedright\arraybackslash}X>{\hsize=1.0000\hsize\linewidth=\hsize\raggedright\arraybackslash}X>{\hsize=1.0000\hsize\linewidth=\hsize\raggedright\arraybackslash}X>{\hsize=1.0000\hsize\linewidth=\hsize\raggedright\arraybackslash}X>{\hsize=1.0000\hsize\linewidth=\hsize\raggedright\arraybackslash}X@{}}
\toprule
\textbf{Dataset} & \textbf{Full prefill NDCG@10} & \textbf{\ensuremath{\Delta} NDCG@10} & \textbf{\ensuremath{\Delta} NDCG@5} & \textbf{\ensuremath{\Delta} HR@1} & \textbf{\ensuremath{\Delta} HR@5} & \textbf{Naive KV reuse: \ensuremath{\Delta} NDCG@10} \\
\midrule
Amazon Books & 0.621 & +0.008 [\ensuremath{-}0.015, +0.030] & +0.009 [\ensuremath{-}0.022, +0.040] & +0.015 [\ensuremath{-}0.024, +0.053] & +0.001 [\ensuremath{-}0.037, +0.039] & \ensuremath{-}0.164 (\ensuremath{-}26.4\%) \\
Goodreads & 0.617 & \ensuremath{-}0.001 [\ensuremath{-}0.011, +0.009] & +0.001 [\ensuremath{-}0.012, +0.014] & \ensuremath{-}0.001 [\ensuremath{-}0.020, +0.017] & +0.005 [\ensuremath{-}0.010, +0.019] & \ensuremath{-}0.164 (\ensuremath{-}26.6\%) \\
Amazon Movies \& TV & 0.659 & +0.006 [\ensuremath{-}0.006, +0.018] & +0.011 [\ensuremath{-}0.005, +0.027] & +0.001 [\ensuremath{-}0.021, +0.022] & +0.016 [\ensuremath{-}0.001, +0.033] & \ensuremath{-}0.205 (\ensuremath{-}31.1\%) \\
Yelp & 0.665 & +0.008 [\ensuremath{-}0.006, +0.022] & +0.012 [\ensuremath{-}0.005, +0.031] & +0.010 [\ensuremath{-}0.015, +0.037] & +0.014 [\ensuremath{-}0.003, +0.033] & \ensuremath{-}0.210 (\ensuremath{-}31.6\%) \\
\bottomrule
\end{tabularx}
\endgroup
\end{minipage}
\par\addvspace{10pt}

\textbf{First, KVUse maintains the ranking quality of full prefill, whereas naive KV reuse loses a large part of it.} In every round on Amazon Books, the NDCG@10 difference between KVUse and full prefill lies between \ensuremath{-}0.0195 and +0.0246, and none of these differences is statistically significant (all \emph{p} \textgreater{} 0.10). The same holds on all four datasets and under all four metrics: averaged over the rounds, KVUse changes NDCG@10 by \ensuremath{-}0.001 to +0.008, none of the differences in Table 4 is statistically significant, and no dataset shows a significant NDCG@10 difference in any single round. Naive KV reuse, which reads the same stored KV but only corrects its positions, lowers NDCG@10 by 26\%--27\% in every round on Amazon Books and by 27\%--32\% on the other three datasets.

\textbf{Second, errors do not accumulate along the memory chain.} KVUse's only shortfall is in round 0, where it scores 0.0195 below full prefill, a difference that is not significant (\emph{p} = 0.153). From round 1 on, as each update call folds the user's new interactions into the profile, the difference turns neutral and slightly positive (not statistically significant). NDCG@5 and HR@1 follow the same pattern, and HR@5 changes by less than 0.02 in every round (Table 8). Goodreads and Yelp follow the same pattern at a smaller scale (\ensuremath{-}0.002 and \ensuremath{-}0.001 in round 0, +0.002 and +0.014 in round 3).

\subsection{Profile Fidelity}

Beyond downstream ranking quality, we examine whether reusing KV changes how accurately the user profile describes the user's interests and engagement habits. In KVUse, the user profile is written by an update call that reads reused KV of previous profile and decode with the writer on, so it may state the user's preferences differently from the profile that the vanilla LLM writes with full prefill, even when both lead to the same ranking.

We therefore compare the two profiles (e.g. user profiles in KVUse and full prefill) on Amazon Books with an LLM judge, Gemini-3-Flash. For each user and profile update round, the judge receives the full-prefill profile and the KVUse profile and judges whether the two convey the same meaning, that is, the same core preferences of the user. Following its instructions, it ignores minor details and answers with one of five levels:

\begin{itemize}

\item \textbf{Equivalent (5):} both profiles state the same main preferences at the same level of specificity; they differ at most in wording, order, or detail.

\item \textbf{Minor (4):} the profiles differ in a single peripheral detail or in emphasis, which would not change the main preferences.

\item \textbf{Partial (3):} the main preferences overlap, but each profile includes a preference missing from the other, or one profile is substantially narrower.

\item \textbf{Mostly different (2):} the profiles share at most a small part of their preferences and would lead to different rankings.

\item \textbf{Unrelated (1):} the profiles have no meaningful preference in common.

\end{itemize}

Table 5 reports the verdicts. KVUse preserves the content of the profile: 79.8\% of its profiles are judged equivalent to the full-prefill profile and a further 14.5\% differ only in a minor detail, so 94.4\% (604 of 640). The remaining 5.6\% are partial, and no profile is judged mostly different or unrelated. The verdicts also show that differences do not compound along the memory chain. The mean level is 4.97 in round 0, where the initialization call reads no earlier profile KV and then holds around 4.7 in following rounds. Together with Section 4.2, these results show that KVUse preserves both the downstream ranking and the content fidelity of the user memory.

\par\addvspace{12pt}
\noindent\begin{minipage}{\linewidth}
\centering
\captionof{table}{KVUse versus full-prefill profile verdicts (640 user-rounds) and controls. Mean: 1--5 scale; percentages: levels 5--1.}\label{tab:5}
\begingroup
\fontsize{7.0}{8.8}\selectfont
\setlength{\tabcolsep}{3pt}
\renewcommand{\arraystretch}{1.18}
\begin{tabularx}{\linewidth}{@{}>{\hsize=1.0000\hsize\linewidth=\hsize\raggedright\arraybackslash}X>{\hsize=1.0000\hsize\linewidth=\hsize\raggedright\arraybackslash}X>{\hsize=1.0000\hsize\linewidth=\hsize\raggedright\arraybackslash}X>{\hsize=1.0000\hsize\linewidth=\hsize\raggedright\arraybackslash}X>{\hsize=1.0000\hsize\linewidth=\hsize\raggedright\arraybackslash}X>{\hsize=1.0000\hsize\linewidth=\hsize\raggedright\arraybackslash}X>{\hsize=1.0000\hsize\linewidth=\hsize\raggedright\arraybackslash}X@{}}
\toprule
\textbf{} & \textbf{Mean} & \textbf{5} & \textbf{4} & \textbf{3} & \textbf{2} & \textbf{1} \\
\midrule
Round 0 & 4.97 & 96.9\% & 3.1\% & 0.0\% & 0.0\% & 0.0\% \\
Round 1 & 4.70 & 76.2\% & 17.5\% & 6.2\% & 0.0\% & 0.0\% \\
Round 2 & 4.64 & 72.5\% & 18.8\% & 8.8\% & 0.0\% & 0.0\% \\
Round 3 & 4.66 & 73.8\% & 18.8\% & 7.5\% & 0.0\% & 0.0\% \\
All rounds & 4.74 & 79.8\% & 14.5\% & 5.6\% & 0.0\% & 0.0\% \\
Control: different users & 1.37 & 0.0\% & 0.0\% & 6.7\% & 23.3\% & 70.0\% \\
Control: same profile & 5.00 & 100\% & 0\% & 0\% & 0\% & 0\% \\
\bottomrule
\end{tabularx}
\endgroup
\end{minipage}
\par\addvspace{10pt}

\subsection{Serving Throughput and Latency}

We next measure how much KVUse improves the serving throughputs of the two paths. Table 6 reports the saturated throughput of KVUse and full prefill on the four datasets. We first discuss the two paths on Amazon Books, starting with the downstream scoring path, which carries most of the traffic and is latency sensitive (Table 1), and then turn to the other datasets.

On the downstream scoring path, KVUse raises throughput by 3.96\ensuremath{\times}, from 31.99 to 126.72 users per second, and shortens the mean engine step time by 3.98\ensuremath{\times}, from 642 ms to 161 ms. This large speedup all comes from the saved redundant computation during the prefill stage, which takes about 86\% of the engine work of the full prefill baseline (Table 1). KVUse instead serves the user profile and the item row of every scoring prompt from the KV storage and computes only the few tokens that they do not cover. Prefix caching cannot recover this work for the baseline, because the item row and the instruction follow a profile whose length varies from user to user, so they never form a shared prefix.

On the profile update path, KVUse raises throughput by 1.61\ensuremath{\times} in round 0, from 54.9 to 88.4 requests per second, and shortens the mean engine step time by 1.57\ensuremath{\times}, from 100 ms to 64 ms. The gain is smaller than on the scoring path because the update call spends significant time on decoding (180 tokens on average in round 0) while the scoring path spends majority of the latency on prefill, where the KV reuse can speed up.

The gains carry over to the other three datasets. On Goodreads, Amazon Movies \& TV, and Yelp, KVUse raises scoring-path throughput by 4.06\ensuremath{\times}--4.49\ensuremath{\times}. On the update path, its gain is 1.45\ensuremath{\times}--1.91\ensuremath{\times} and falls from round to round, because the profile, and with it the decode, grows from 160--169 output tokens in round 0 to 237--306 in round 3.

\par\addvspace{12pt}
\noindent\begin{minipage}{\linewidth}
\centering
\captionof{table}{Saturated scoring: complete users/s (speedup); update: rounds 0--3. Speedups: KVUse/full prefill.}\label{tab:6}
\begingroup
\fontsize{8.8}{11.0}\selectfont
\setlength{\tabcolsep}{3pt}
\renewcommand{\arraystretch}{1.18}
\begin{tabularx}{\linewidth}{@{}>{\hsize=1.0000\hsize\linewidth=\hsize\raggedright\arraybackslash}X>{\hsize=1.0000\hsize\linewidth=\hsize\raggedright\arraybackslash}X>{\hsize=1.0000\hsize\linewidth=\hsize\raggedright\arraybackslash}X>{\hsize=1.0000\hsize\linewidth=\hsize\raggedright\arraybackslash}X@{}}
\toprule
\textbf{Dataset} & \textbf{Scoring: full prefill} & \textbf{Scoring: KVUse} & \textbf{Update: rounds 0 / 1 / 2 / 3} \\
\midrule
Amazon Books & 31.99 & 126.72 (3.96\ensuremath{\times}) & 1.61\ensuremath{\times} / 1.55\ensuremath{\times} / 1.48\ensuremath{\times} / 1.43\ensuremath{\times} \\
Goodreads & 29.44 & 125.44 (4.26\ensuremath{\times}) & 1.77\ensuremath{\times} / 1.58\ensuremath{\times} / 1.52\ensuremath{\times} / 1.45\ensuremath{\times} \\
Amazon Movies \& TV & 31.18 & 126.72 (4.06\ensuremath{\times}) & 1.60\ensuremath{\times} / 1.57\ensuremath{\times} / 1.53\ensuremath{\times} / 1.48\ensuremath{\times} \\
Yelp & 26.23 & 117.76 (4.49\ensuremath{\times}) & 1.91\ensuremath{\times} / 1.72\ensuremath{\times} / 1.61\ensuremath{\times} / 1.58\ensuremath{\times} \\
\bottomrule
\end{tabularx}
\endgroup
\end{minipage}
\par\addvspace{10pt}

\subsection{Ablation Studies}

\subsubsection{Hiding the KV Upload}

We compare KVUse with a variant that copies the stored KV from host memory synchronously, on the critical path of each engine step. On Amazon Books, hiding the upload raises the scoring-path throughput gain from 3.40\ensuremath{\times} to 3.96\ensuremath{\times} (108.80 versus 126.72 users per second). Without the KV scheduler, the stored KV is copied from host memory on the critical path, which stalls computation between adjacent batches. On the update path, the two variants differ little (1.41\ensuremath{\times}--1.57\ensuremath{\times} versus 1.43\ensuremath{\times}--1.61\ensuremath{\times}), because uploading the KV takes little time compared with decoding the profile. On all four datasets, hiding the upload raises the scoring-path gain by a further 1.16\ensuremath{\times}--1.20\ensuremath{\times} (Table 7).

\par\addvspace{12pt}
\noindent\begin{minipage}{\linewidth}
\centering
\captionof{table}{Effect of hiding KV upload. Speedups over full prefill; update ranges cover all four memory rounds.}\label{tab:7}
\begingroup
\fontsize{8.2}{10.2}\selectfont
\setlength{\tabcolsep}{3pt}
\renewcommand{\arraystretch}{1.18}
\begin{tabularx}{\linewidth}{@{}>{\hsize=1.0000\hsize\linewidth=\hsize\raggedright\arraybackslash}X>{\hsize=1.0000\hsize\linewidth=\hsize\raggedright\arraybackslash}X>{\hsize=1.0000\hsize\linewidth=\hsize\raggedright\arraybackslash}X>{\hsize=1.0000\hsize\linewidth=\hsize\raggedright\arraybackslash}X>{\hsize=1.0000\hsize\linewidth=\hsize\raggedright\arraybackslash}X@{}}
\toprule
\textbf{Dataset} & \textbf{Scoring: not hidden} & \textbf{Scoring: hidden} & \textbf{Update: not hidden} & \textbf{Update: hidden} \\
\midrule
Amazon Books & 3.40\ensuremath{\times} & 3.96\ensuremath{\times} & 1.41\ensuremath{\times}--1.57\ensuremath{\times} & 1.43\ensuremath{\times}--1.61\ensuremath{\times} \\
Goodreads & 3.57\ensuremath{\times} & 4.26\ensuremath{\times} & 1.40\ensuremath{\times}--1.74\ensuremath{\times} & 1.45\ensuremath{\times}--1.77\ensuremath{\times} \\
Amazon Movies \& TV & 3.49\ensuremath{\times} & 4.06\ensuremath{\times} & 1.36\ensuremath{\times}--1.58\ensuremath{\times} & 1.48\ensuremath{\times}--1.60\ensuremath{\times} \\
Yelp & 3.76\ensuremath{\times} & 4.49\ensuremath{\times} & 1.54\ensuremath{\times}--1.81\ensuremath{\times} & 1.58\ensuremath{\times}--1.91\ensuremath{\times} \\
\bottomrule
\end{tabularx}
\endgroup
\end{minipage}
\par\addvspace{10pt}

\subsubsection{Writer Transfer Across Datasets}

To test whether writers must be trained for each dataset, we apply the Amazon Books writers unchanged to the other three datasets (Table 9 in Appendix B). They transfer to Amazon Movies \& TV, where NDCG@10 stays at the level of full prefill (+0.007), but not to Goodreads (\ensuremath{-}0.022) or Yelp (\ensuremath{-}0.137), on which in-domain writers show no loss. For now, writers must therefore be trained for each dataset (Appendix C).

\FloatBarrier

\section{Conclusion}

This paper presents KVUse, a memory-augmented LLM recommender that stores each user's profile as reusable KV instead of text. LoRA writers, trained by self-distillation from the vanilla LLM, let the update call produce the KV of the new profile while decoding it, and let the frozen LLM read this KV, together with precomputed item KV, at new positions and behind new context without recomputation. A KV scheduler uploads the stored KV of the next batch while the current batch is computed, hiding the upload behind GPU computation. On four InstructRec datasets and over four memory updating rounds, KVUse maintains the ranking quality of full prefill under NDCG@10, NDCG@5, HR@1, and HR@5, whereas naive KV reuse loses 26\%--32\% of NDCG@10. Under saturated load, KVUse raises the throughput of the downstream scoring path by 4.0\ensuremath{\times}--4.5\ensuremath{\times} and of the profile update path by 1.4\ensuremath{\times}--1.9\ensuremath{\times}.

\FloatBarrier

\FloatBarrier

\clearpage

\section*{Appendix A: Prompt Examples}

Figures 4 and 5 show example prompts and responses of the two calls of the memory loop (Section 3.1). The user profile is shown in red and the item descriptions in orange; these are the spans whose KV KVUse stores and reuses.

\begin{figure}[!htbp]
\centering
\begin{minipage}[t]{0.67\linewidth}
\vspace{0pt}
\begin{tcolorbox}[colback=metabg,colframe=metabg,boxrule=0pt,arc=2pt,left=5pt,right=5pt,top=5pt,bottom=5pt]
{\sffamily\bfseries Prompt}\par\smallskip
\fontsize{8.8}{10.8}\selectfont\ttfamily
\setlength{\parskip}{2pt}\setlength{\parindent}{0pt}
\hyphenpenalty=0\exhyphenpenalty=0
{\color[HTML]{37474F}You are a user preference analyst. Write concise, coherent profiles.}
\par

\par
{\color[HTML]{37474F}Below is a user's current preference profile (empty if this is the 1st time we generate the user profile of this user), followed by the most recent items they engaged with since that profile was written. Write a single coherent paragraph (150-300 words) in third person, beginning with "This user". The profile should account for both the new activity and previous profile (if it is non-empty). Do NOT use bullet points, and do not mention the update, the previous profile, or the passage of time.}
\par

\par
{\color[HTML]{37474F}\#\# Current profile}
\par
\textbf{{\color[HTML]{FF0000}This user is drawn to a diverse range of topics, from self help and personal development to historical non fiction and fiction...}}
\par

\par
{\color[HTML]{37474F}\#\# New activity
\par
| ID | Title | Description |
\par
|-----|-------|-------------|}
\par
{\bfseries\sffamily {\color[HTML]{FF9900}| book\_123531 | Justice for Speedy: A True Horse Story | Justice for Speedy: A True Horse Story. Judy Berkley has been a published writer since 1984. In addition to her freelance work, Berkley was a daily columnist at the Oakland Tribune newspaper in northern |
\par
| book\_180352 | Gatefather: A Novel of the Mithermages | Gatefather: A Novel of the Mithermages. This ambitious tale is well crafted, highly detailed, and pleasantly accessible. \textless{}i\textgreater{}Publishers Weekly on The Lost Gate\textless{}/i\textgreater{} |}}
\par

\par
{\color[HTML]{37474F}Generate the output now:}
\end{tcolorbox}
\end{minipage}\hfill
\begin{minipage}[t]{0.3\linewidth}
\vspace{0pt}
\begin{tcolorbox}[colback=metabg,colframe=metabg,boxrule=0pt,arc=2pt,left=5pt,right=5pt,top=5pt,bottom=5pt]
{\sffamily\bfseries Output}\par\smallskip
\fontsize{8.8}{10.8}\selectfont\ttfamily
\setlength{\parskip}{2pt}\setlength{\parindent}{0pt}
\hyphenpenalty=0\exhyphenpenalty=0
{\color[HTML]{FF0000}\textbf{This user is drawn to a diverse range of topics, from self-help and personal development to historical non-fiction and fiction. Their interest in "You're Not Crazy - It's Your Mother" suggests a desire to understand and cope with complex family dynamics ...}}
\end{tcolorbox}
\end{minipage}
\caption{An example LLM prompt (left) and response (right) of the user profile update path. For space, the current profile and item descriptions are truncated.}\label{fig:4}
\end{figure}

\begin{figure}[!htbp]
\centering
\begin{minipage}[t]{0.67\linewidth}
\vspace{0pt}
\begin{tcolorbox}[colback=metabg,colframe=metabg,boxrule=0pt,arc=2pt,left=5pt,right=5pt,top=5pt,bottom=5pt]
{\sffamily\bfseries Prompt}\par\smallskip
\fontsize{8.8}{10.8}\selectfont\ttfamily
\setlength{\parskip}{2pt}\setlength{\parindent}{0pt}
\hyphenpenalty=0\exhyphenpenalty=0
{\color[HTML]{37474F}You are a recommendation engine. Score each item for this user.}
\par

\par
{\color[HTML]{37474F}\#\# Instructions
\par
Score each item on how well it matches the user's preferences (0-10 scale):
\par
- 8-10: STRONG match --- high confidence user will enjoy this
\par
- 4-7: MAYBE --- moderate match, could work
\par
- 0-3: WEAK --- poor match for this user}
\par

\par
{\color[HTML]{37474F}Consider:
\par
- How well does the item match the user's stated preferences?
\par
- Does it align with their likes or trigger their dislikes?
\par
- Is it a good fit for their demonstrated taste?}
\par

\par
{\color[HTML]{37474F}\#\# Output Format
\par
One line per item, no extra text:
\par
SCORE}
\par

\par
{\color[HTML]{37474F}Example:
\par
8.5}
\par

\par
{\color[HTML]{37474F}Below are the given User Preferences and Items}
\par

\par
{\color[HTML]{37474F}\#\# User Preferences}
\par
\textbf{{\color[HTML]{FF0000}This user is drawn to a diverse range of topics, from self-help and personal development to historical non-fiction and fiction. Their interest in "You're Not Crazy - It's Your Mother" suggests a desire to understand and cope with complex family dynamics ...}}
\par

\par
{\color[HTML]{37474F}\#\# Items to Score
\par
| ID | Title | Description |
\par
|-----|-------|-------------|}
\par
\textbf{{\color[HTML]{FF9900}| book\_74982 | Taka (La' Femme Fatale' Publishing) | Taka (La' Femme Fatale' Publishing). Author Tyberia Blaqk is a mother of three who began writing stories as early as the 4th grade when she discovered that she had a real talent for bringing fictional |}}
\par

\par
{\color[HTML]{37474F}Generate the output now:}
\end{tcolorbox}
\end{minipage}\hfill
\begin{minipage}[t]{0.3\linewidth}
\vspace{0pt}
\begin{tcolorbox}[colback=metabg,colframe=metabg,boxrule=0pt,arc=2pt,left=5pt,right=5pt,top=5pt,bottom=5pt]
{\sffamily\bfseries Output}\par\smallskip
\fontsize{8.8}{10.8}\selectfont\ttfamily
\setlength{\parskip}{2pt}\setlength{\parindent}{0pt}
\hyphenpenalty=0\exhyphenpenalty=0
{\color[HTML]{37474F}6.2}
\end{tcolorbox}
\end{minipage}
\caption{An example LLM prompt (left) and response (right) of the downstream scoring path. For space, the current profile and item descriptions are truncated.}\label{fig:5}
\end{figure}

\FloatBarrier

\clearpage

\section*{Appendix B: Additional Results}

Table 8 gives all four ranking metrics per memory round on Amazon Books (Section 4.2), Table 9 gives the writer-transfer results of Section 4.5.2, and Section B.1 compares KVUse with read-time repair (Table 10).

\par\addvspace{12pt}
\noindent\begin{minipage}{\linewidth}
\centering
\captionof{table}{Ranking quality per memory update round on Amazon Books, 160 held-out users. For each metric, the first three rows give full prefill, naive KV reuse (relocation only), and KVUse, and the \ensuremath{\Delta} row gives KVUse minus full prefill.}\label{tab:8}
\begingroup
\fontsize{8.2}{10.2}\selectfont
\setlength{\tabcolsep}{3pt}
\renewcommand{\arraystretch}{1.18}
\begin{tabularx}{\linewidth}{@{}>{\hsize=0.9500\hsize\linewidth=\hsize\raggedright\arraybackslash}X>{\hsize=1.8500\hsize\linewidth=\hsize\raggedright\arraybackslash}X>{\hsize=0.8000\hsize\linewidth=\hsize\raggedright\arraybackslash}X>{\hsize=0.8000\hsize\linewidth=\hsize\raggedright\arraybackslash}X>{\hsize=0.8000\hsize\linewidth=\hsize\raggedright\arraybackslash}X>{\hsize=0.8000\hsize\linewidth=\hsize\raggedright\arraybackslash}X@{}}
\toprule
\textbf{Metric} & \textbf{Arm} & \textbf{Round 0} & \textbf{Round 1} & \textbf{Round 2} & \textbf{Round 3} \\
\midrule
NDCG@10 & Full prefill & 0.6152 & 0.6128 & 0.6306 & 0.6255 \\
 & Naive KV reuse & 0.4564 & 0.4491 & 0.4658 & 0.4559 \\
 & \textbf{KVUse}  & \textbf{0.5957} & \textbf{0.6248} & \textbf{0.6454} & \textbf{0.6502} \\
 & \ensuremath{\Delta} (KVUse \ensuremath{-} full) & \ensuremath{-}0.0195 & +0.0119 & +0.0148 & +0.0246 \\
\addlinespace
NDCG@5 & Full prefill & 0.5245 & 0.5220 & 0.5442 & 0.5368 \\
 & Naive KV reuse & 0.3007 & 0.2868 & 0.3124 & 0.2972 \\
 & \textbf{KVUse}  & \textbf{0.5001} & \textbf{0.5347} & \textbf{0.5608} & \textbf{0.5678} \\
 & \ensuremath{\Delta} (KVUse \ensuremath{-} full) & \ensuremath{-}0.0244 & +0.0127 & +0.0166 & +0.0311 \\
\addlinespace
HR@1 & Full prefill & 0.3323 & 0.3121 & 0.3529 & 0.3296 \\
 & Naive KV reuse & 0.1003 & 0.0938 & 0.1133 & 0.1004 \\
 & \textbf{KVUse}  & \textbf{0.2993} & \textbf{0.3409} & \textbf{0.3744} & \textbf{0.3713} \\
 & \ensuremath{\Delta} (KVUse \ensuremath{-} full) & \ensuremath{-}0.0330 & +0.0288 & +0.0215 & +0.0417 \\
\addlinespace
HR@5 & Full prefill & 0.7147 & 0.7140 & 0.7265 & 0.7220 \\
 & Naive KV reuse & 0.5120 & 0.4903 & 0.5187 & 0.5029 \\
 & \textbf{KVUse}  & \textbf{0.6969} & \textbf{0.7129} & \textbf{0.7321} & \textbf{0.7393} \\
 & \ensuremath{\Delta} (KVUse \ensuremath{-} full) & \ensuremath{-}0.0178 & \ensuremath{-}0.0012 & +0.0056 & +0.0174 \\
\bottomrule
\end{tabularx}
\endgroup
\end{minipage}
\par\addvspace{10pt}

\par\addvspace{12pt}
\noindent\begin{minipage}{\linewidth}
\centering
\captionof{table}{NDCG@10 change relative to full prefill when the Amazon Books writers are applied unchanged to the other three datasets, averaged over the four memory rounds, with 95\% confidence intervals over users in brackets. In-domain writers are shown for reference (Table 4).}\label{tab:9}
\begingroup
\fontsize{8.8}{11.0}\selectfont
\setlength{\tabcolsep}{3pt}
\renewcommand{\arraystretch}{1.18}
\begin{tabularx}{\linewidth}{@{}>{\hsize=1.0000\hsize\linewidth=\hsize\raggedright\arraybackslash}X>{\hsize=1.0000\hsize\linewidth=\hsize\raggedright\arraybackslash}X>{\hsize=1.0000\hsize\linewidth=\hsize\raggedright\arraybackslash}X@{}}
\toprule
\textbf{Dataset} & \textbf{In-domain writers} & \textbf{Amazon Books writers} \\
\midrule
Goodreads & \ensuremath{-}0.001 [\ensuremath{-}0.011, +0.009] & \textbf{\ensuremath{-}0.022} [\ensuremath{-}0.034, \ensuremath{-}0.010] \\
Amazon Movies \& TV & +0.006 [\ensuremath{-}0.006, +0.018] & +0.007 [\ensuremath{-}0.006, +0.020] \\
Yelp & +0.008 [\ensuremath{-}0.006, +0.022] & \textbf{\ensuremath{-}0.137} [\ensuremath{-}0.157, \ensuremath{-}0.118] \\
\bottomrule
\end{tabularx}
\endgroup
\end{minipage}
\par\addvspace{10pt}

\subsection*{B.1. Comparison with Read-Time Repair}

KVUse repairs the reused KV when it writes it. Two training-free methods, CacheBlend \citep{ref24} and EPIC \citep{ref8}, instead repair it when they read it, by recomputing part of the reused tokens in every request (Section 2.2). We replace the writers of KVUse on Amazon Books with each of these two methods and keep everything else unchanged. Both methods read the KV that the vanilla LLM computes for each span on its own, re-rotated to its position, with a recomputation budget of about 15\% of the reused tokens: CacheBlend-15 recomputes 15\% of them, and EPIC with LegoLink-11 recomputes 15.9\% in the scoring prompts and 18.4\% in the update prompts.

At this budget, read-time repair recovers almost none of the loss (Table 10). Averaged over the four rounds, CacheBlend-15 and EPIC lower NDCG@10 by 0.162 and 0.164, nearly as much as naive KV reuse (0.166), and their losses are statistically significant in every round and under every metric, whereas KVUse, which recomputes nothing at read time, shows no significant difference from full prefill under any metric. Read-time repair closes the gap only with a much larger budget: CacheBlend reaches parity when it recomputes 50\% of the reused tokens (\ensuremath{-}0.006 NDCG@10), while EPIC still loses 0.052 when it recomputes 32\%. Write-time repair therefore removes the context error without adding computation to any request.

\par\addvspace{12pt}
\noindent\begin{minipage}{\linewidth}
\centering
\captionof{table}{Ranking quality of write-time repair (KVUse) and read-time repair (CacheBlend, EPIC) at a budget of about 15\% recomputed tokens on Amazon Books, as the change relative to full prefill averaged over the four memory rounds, with 95\% confidence intervals over users in brackets. The second column gives the share of reused tokens recomputed in the scoring and update prompts. These runs use all 200 evaluation users.}\label{tab:10}
\begingroup
\fontsize{7.5}{9.4}\selectfont
\setlength{\tabcolsep}{3pt}
\renewcommand{\arraystretch}{1.18}
\begin{tabularx}{\linewidth}{@{}>{\hsize=1.0000\hsize\linewidth=\hsize\raggedright\arraybackslash}X>{\hsize=1.0000\hsize\linewidth=\hsize\raggedright\arraybackslash}X>{\hsize=1.0000\hsize\linewidth=\hsize\raggedright\arraybackslash}X>{\hsize=1.0000\hsize\linewidth=\hsize\raggedright\arraybackslash}X>{\hsize=1.0000\hsize\linewidth=\hsize\raggedright\arraybackslash}X>{\hsize=1.0000\hsize\linewidth=\hsize\raggedright\arraybackslash}X@{}}
\toprule
\textbf{Method} & \textbf{Reused tokens recomputed (scoring / update)} & \textbf{\ensuremath{\Delta} NDCG@10} & \textbf{\ensuremath{\Delta} NDCG@5} & \textbf{\ensuremath{\Delta} HR@1} & \textbf{\ensuremath{\Delta} HR@5} \\
\midrule
KVUse & 0\% / 0\% & +0.005 [\ensuremath{-}0.015, +0.025] & +0.006 [\ensuremath{-}0.022, +0.034] & +0.006 [\ensuremath{-}0.028, +0.040] & +0.001 [\ensuremath{-}0.034, +0.035] \\
CacheBlend & 15.0\% / 15.0\% & \ensuremath{-}0.162 [\ensuremath{-}0.196, \ensuremath{-}0.129] & \ensuremath{-}0.228 [\ensuremath{-}0.276, \ensuremath{-}0.181] & \ensuremath{-}0.231 [\ensuremath{-}0.285, \ensuremath{-}0.179] & \ensuremath{-}0.204 [\ensuremath{-}0.259, \ensuremath{-}0.148] \\
EPIC & 15.9\% / 18.4\% & \ensuremath{-}0.164 [\ensuremath{-}0.200, \ensuremath{-}0.129] & \ensuremath{-}0.235 [\ensuremath{-}0.284, \ensuremath{-}0.185] & \ensuremath{-}0.230 [\ensuremath{-}0.285, \ensuremath{-}0.175] & \ensuremath{-}0.221 [\ensuremath{-}0.271, \ensuremath{-}0.168] \\
Naive KV reuse & 0\% / 0\% & \ensuremath{-}0.166 [\ensuremath{-}0.201, \ensuremath{-}0.131] & \ensuremath{-}0.237 [\ensuremath{-}0.285, \ensuremath{-}0.188] & \ensuremath{-}0.232 [\ensuremath{-}0.288, \ensuremath{-}0.178] & \ensuremath{-}0.222 [\ensuremath{-}0.270, \ensuremath{-}0.171] \\
\bottomrule
\end{tabularx}
\endgroup
\end{minipage}
\par\addvspace{10pt}

\FloatBarrier

\clearpage

\section*{Appendix C: Limitations}

\textbf{Transfer across domains.} The writers are trained for each dataset. Writers trained on Amazon Books transfer to Amazon Movies \& TV but not to Goodreads or Yelp (Section 4.5.2), so a new domain currently needs its own training data and writers. Training writers that generalize across domains, for example on a mixture of datasets, is left to future work.

\textbf{Upload hiding requires knowing the next batch.} The KV scheduler can upload the KV of the next batch while the current batch is computed only if the next requests are known in advance. This holds in offline and near-line jobs, such as batch scoring and the asynchronous profile updates, whose requests are queued before they are served. When requests arrive online and the next batch cannot be known in advance, KVUse still avoids recomputing the stored KV, but it can no longer fully hide the upload behind computation; on Amazon Books, the scoring-path gain without hiding is 3.40\ensuremath{\times} rather than 3.96\ensuremath{\times} (Section 4.5.1).

\textbf{Storage.} KVUse trades storage for computation. Text takes a few bytes per token, whereas the KV of Llama-3.1-8B-Instruct takes 128 KiB per token in bfloat16, so the KV storage of KVUse is several orders of magnitude larger than the text that the full-prefill baseline stores. Compressing the stored KV, for example by quantizing it or by learning to compress it into a compact memory \citep{ref9}, could reduce this storage needs and is a direction for future work.

\FloatBarrier

\clearpage

\section*{Appendix D: Training Recipe and Gates}

Figure 6 illustrates the training of one sample (Section 3.3), and Table 11 lists the complete settings. Four runtime gates check that each run trains what it claims: (1) a gradient probe confirms that exactly the claimed writers receive gradient; (2) a per-loss probe records which writers each loss reaches; (3) a seeding check refuses a Stage B seed that was trained on the wrong loss set; and (4) a post-training check verifies that every declared writer has non-zero weights.

\begin{figure}[!htbp]
\centering
\includegraphics[width=\linewidth]{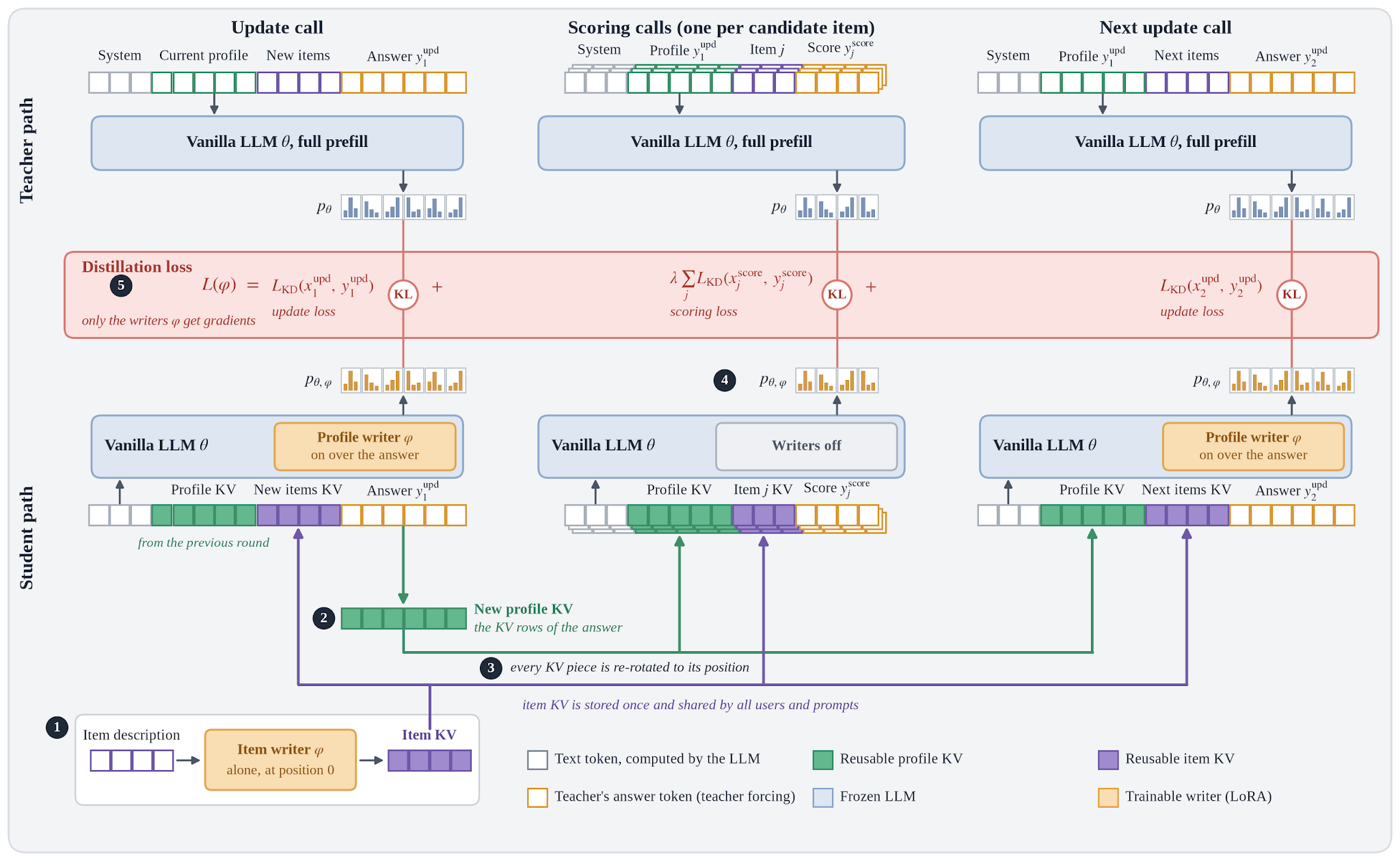}
\caption{KVUse training for one sample, which consists of an update call, the scoring calls that read the profile this call writes, and the next update call. Top: the teacher path, in which the vanilla LLM reads every complete prompt with full prefill. Bottom: the student path, in which the profile and item spans are supplied as KV written by the writers; circled numbers 1--5 mark the five steps described in the text. Middle: the distillation loss of Eq. (2), which compares the next-token distributions of the two paths at every answer token.}\label{fig:6}
\end{figure}

\par\addvspace{12pt}
\noindent\begin{minipage}{\linewidth}
\centering
\captionof{table}{Complete training settings.}\label{tab:11}
\begingroup
\fontsize{8.8}{11.0}\selectfont
\setlength{\tabcolsep}{3pt}
\renewcommand{\arraystretch}{1.18}
\begin{tabularx}{\linewidth}{@{}>{\hsize=0.5500\hsize\linewidth=\hsize\raggedright\arraybackslash}X>{\hsize=1.4500\hsize\linewidth=\hsize\raggedright\arraybackslash}X@{}}
\toprule
\textbf{Item} & \textbf{Value} \\
\midrule
Base model (frozen reader) & Llama-3.1-8B-Instruct \\
Writer & LoRA on q/k/v/o, rank 8, alpha 16, 6.8M trainable parameters each \\
Writers (deployed) & Two: user profile and item descriptions \\
Distillation & KL to the full-prefill teacher, temperature 1.0; targets up to 640 tokens, including end-of-sequence \\
Stage A & Two isolated runs, one per writer, other spans at full prefill; learning rate 5e-4, 2,400-step budget \\
Stage B & Both writers jointly, both losses; learning rate 6e-5, 2,400-step budget, early-stopped at 1,299 steps, best at step 499 \\
Training data & Teacher chains of the 1,500 training users: 4,500 update records and 60,000 scoring records \\
Optimizer & AdamW with 4 training samples per step; linear learning-rate decay to one tenth over the step budget; gradient-norm clipping at 1.0; bfloat16 weights \\
Early stopping & Distillation loss on held-out training users, evaluated every 100 steps, patience 8 \\
Candidates per slate & 10 \\
\bottomrule
\end{tabularx}
\endgroup
\end{minipage}
\par\addvspace{10pt}

\end{document}